\documentclass[review]{elsarticle}
\usepackage{lineno}
\modulolinenumbers[5]
\journal{Computer Vision and Image Understanding}
\pdfoutput=1

\usepackage{graphicx}
\usepackage{subfig}
\usepackage{amsmath,amssymb} % define this before the line numbering.
\usepackage{url}

\usepackage{tabularx}
\usepackage{booktabs}
\usepackage{multirow}

\newcommand{\argmax}{\operatornamewithlimits{arg\,max}}

\newcolumntype{Y}{>{\centering\arraybackslash}X}

\begin{document}

\def\ie{i.e.\ }
\def\eg{e.g.\ }
\def\cf{cf.\ }
\def\etal{et al.\ }

\begin{frontmatter}

\title{Weakly supervised learning of actions from transcripts}

\author{Hilde Kuehne, Alexander Richard, Juergen Gall}
\address{University of Bonn, Germany}




\begin{abstract}
We present an approach for weakly supervised learning of human actions from video transcriptions. Our system is based on the idea that, given a sequence of input data and a transcript, \ie a list of the order the actions occur in the video, it is possible to infer the actions within the video stream and to learn the related action models without the need for any frame-based annotation. 
Starting from the transcript information at hand, we split the given data sequences uniformly based on the number of expected actions. We then learn action models for each class by maximizing the probability that the training video sequences are generated by the action models given the sequence order as defined by the transcripts. The learned model can be used to temporally segment an unseen video with or without transcript. Additionally, the inferred segments can be used as a starting point to train high-level fully supervised models.

We evaluate our approach on four distinct activity datasets, namely Hollywood Extended, MPII Cooking, Breakfast and CRIM13. It shows that the proposed system is able to align the scripted actions with the video data, that the learned models localize and classify actions in the datasets, and that they outperform any current state-of-the-art approach for aligning transcripts with video data. 
\end{abstract}


\end{frontmatter}

%\linenumbers

\section{Introduction}

Weakly supervised learning has become of more and more interest in recent years, also in the field of action recognition (see e.g.~\cite{duchenne2009automatic,sun2015temporal,bojanowski2014weakly,wu2015watch}). Especially the large amount of available data on open video platforms such as Youtube or Vimeo, but also in the context of surveillance, smart-homes, or behavior analysis, constitutes a high demand for weakly supervised methods. 

So far, most methods for action classification and detection rely on supervised learning. This implies that for any new action to be trained, labels with temporal information are needed. But video annotation is very costly and time consuming, as it usually requires not only the label but also corresponding time stamps to capture its relevant content. 
Hence, relying on hand-annotation alone will make it difficult to cover larger amounts of video data and to develop systems to work on large-scale domains outside existing presegmented datasets. Weakly supervised methods, such as the here presented one, might provide a first step towards a remedy in this case. 

\begin{figure}[t]
\begin{center}
\includegraphics[width=0.9\linewidth]{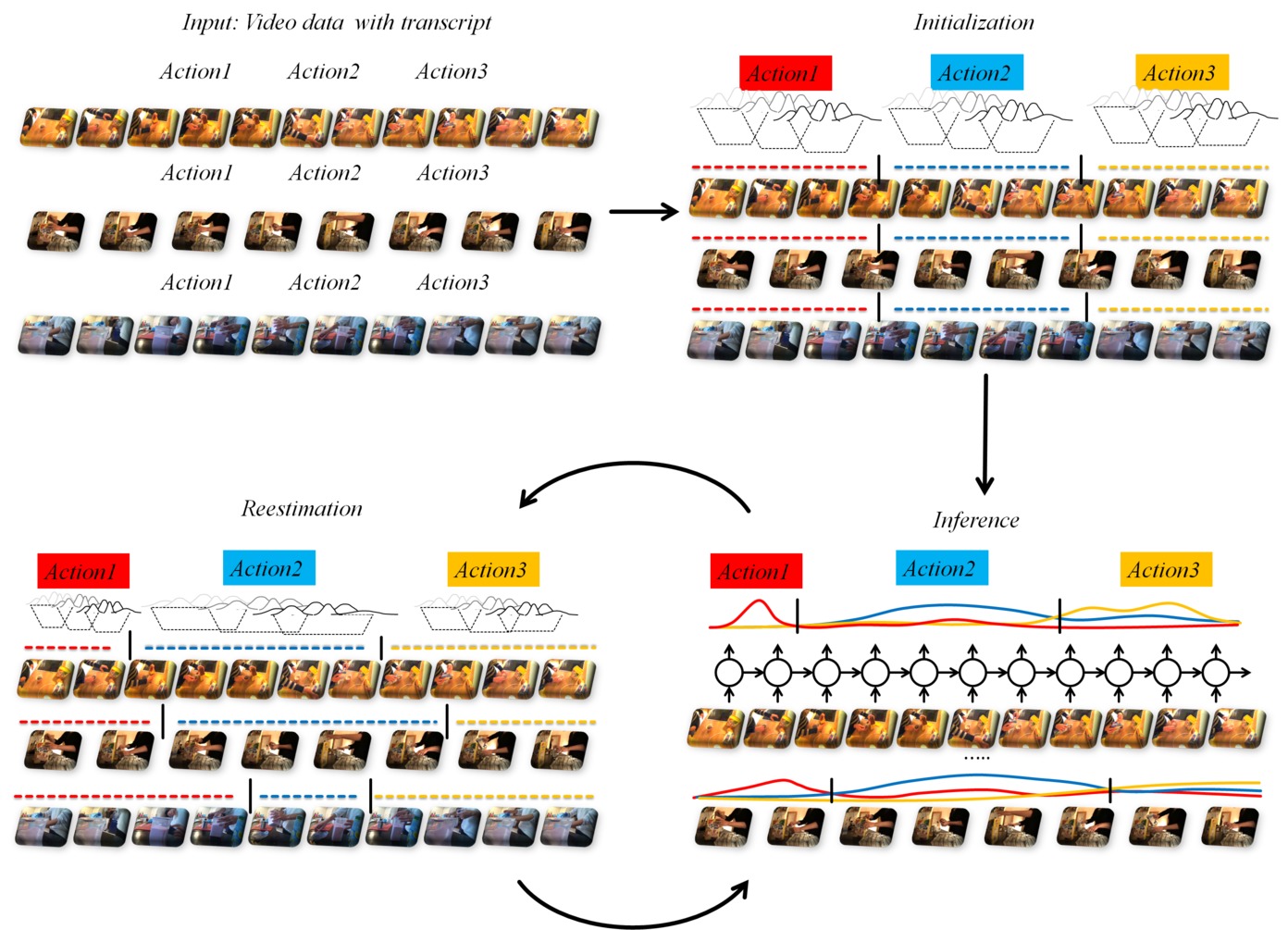} 
\end{center}
\caption{Overview of weakly supervised learning from transcripts. The training data consists of video data and transcripts describing the order of the actions but not providing a labeling at frame level. Each action class is represented by a model that is initialized with uniform segments of the respective videos. The initialized model is used to infer the segments of a video given a transcription of it. The new segmentation is used to reestimate and update the model parameters. The last two steps can be repeated until convergence. }
\label{fig:overview}
\end{figure}

In this work, we propose a framework for weakly supervised learning of temporal action models from video data as shown in Figure \ref{fig:overview}.
The approach is based on a combination of video data and their transcripts as input for training. The transcripts solely describe the order in which the actions occur in the video without the need for a temporal labeling of frames. Based on this input, we lend on the concept of flat models in speech recognition to infer the related actions from the video input. 
We model each action in form of an HMM and learn the model parameters based on the transcripts. First, we initialize each model by generating uniform splits of all videos using the transcript information. During training, we maximize the probability that the training video sequences are generated by the HMMs given the sequence order as defined by the transcripts. Therefore, we infer the segmentation boundaries over all videos and use the updated information to reestimate the model. The observation probabilities are modeled by GMMs, which allow for an efficient, on-the-fly adaptation of the model parameters without retraining. These two steps, inference and update, are repeated multiple times until convergence.

The learned models can then be used to align a transcription of actions to the respective video or to segment or classify a video without any annotation information. 
In the latter case, we learn a grammar from the transcripts of the training data which describes a valid order of actions. The resulting segmentation of the training data can further be used to train any other model in a fully supervised manner and to replace low-level GMM based observation probabilities by ones gained from other models such as CNNs.     

We evaluate our approach on four distinct large-scale datasets for the segmentation of actions in video data, namely the Hollywood Extended \cite{bojanowski2014weakly}, MPII Cooking \cite{rohrbach2012database}, Breakfast \cite{kuehne2014language}, and the CRIM13 \cite{Burgos12CRIM} dataset with overall more than 100h of video data\footnote{Code and data are available under: \\ \url{https://github.com/hildekuehne/WeakActionLearning}}. The proposed system relies on an appropriate amount of data to build robust probabilistic models for initialization and training. 
We compare our results against the state-of-the-art approaches such as \cite{bojanowski2014weakly} and \cite{huang2016connectionist} as well as against a supervised system. It shows that our approach does well compared to the baseline and that it is able to detect actions in unseen sequences based on the weakly trained models. 

%We compare our weakly supervised approach with a fully supervised trained model and show that our approach outperforms the weakly supervised method proposed in~\cite{bojanowski2014weakly}. 

%We evaluate the accuracy of the resulting segmentation as well as the performance of the so trained models for the recognition of new sequences. 

%- complexity: training sequences consist of  (mean sequence length,  mean number of units )
%- runtime: training takes less than .... 
%- Better than supervised annotation (based on time spend)

\section{Related Work}

The problem of action classification on presegmented video clips has been widely studied and considerable advances have been made.
Approaches can be summarized in two different groups, those using hand-crafted generic features and those featuring learned features. Generic features, such as STIPs \cite{laptev2005stip}, dense trajectories \cite{wang2013action}, or improved dense trajectories \cite{wang2013action} are usually used in combination with quantization methods such as Bag-of-words or Fisher vectors. They are based on gradient information, either from the RGB or gray image (HoG) or from the optical flow field (HoF), which is accumulated over a spatial-temporal volume in the video. Especially the combination of Fisher vectors and improved dense trajectories \cite{wang2013action} has shown to give consistently good results on various datasets and is used by various approaches \cite{wang2013action, jain2014action}.
With the advance of deep learning methods, CNN-based approaches are also keeping up with and exceeding the performance of generic features. Here, the most successful configurations, e.g. \cite{wang2016temporal} are usually based on the so called two-stream model \cite{simonyan2014two}. The two stream model consists of two separate CNNs, one for the processing of the RGB image of each frame, one for the processing of the respective optical flow field. The output of both streams can be fused at different network layers, but usually a late fusion is preferred over early or mid-level fusion \cite{karpathy2014large}. Also, many recent approaches use CNNs in combination with improved dense trajectories to reach state-of-the-art results \cite{jain201515,simonyan2014two,feichtenhofer2016conv}. 
In contrast to generic features, CNNs need to be trained. This is, however problematic with regards to weak learning. In this work, we will demonstrate how CNN architectures can be trained with our framework in a weakly supervised context.
%In context of the here proposed work, it is to notice that for learned features, CNNs need to be fine-tuned for each dataset separately to reach state-of-the-art recognition performance. This means that one or more CNNs need to be trained with the training data of a dataset. Opposed to that generic feature extraction and computation is based on fixed algorithms for all datasets and inputs and does not depend on any training information. This characteristic is important for the problem of weak learning as in this case, pre-annotated training data does not exist and, as the here proposed work is based on an iterative approach, CNN fine tuning would be required after each iteration and thus become prohibitively expensive.   

Based on the advances in action classification on presegmented videos, research starts to focus on the task of action detection in unsegmented videos. Here, the aim is to localize and classify a set of actions in temporally unrestricted videos. Some approach the problem as a localization task in the spatiotemporal domain \cite{vanGemert2015apt,jain2014action,yuan2011discriminative} and cluster trajectories or apply dynamic programming to enable efficient search for actions in the input space. Most works, however, focus on finding the temporal location of actions, which can be seen as equivalent to a temporal video segmentation task. E.g. Shi \etal \cite{shi2008discriminative} use a semi-Markov model and find the best segmentation based on support vector machine scores. In \cite{rohrbach2012database}, a sliding window approach with non-maximum suppression is applied and pose features are added to the dense trajectories. Ni \etal \cite{ni2014multiple} analyze the video on multiple levels of granularity and track hands and objects to exploit the most relevant regions for finding actions. In order to adequately model long term dependencies of action segments in videos, a nonparametric hierarchical Bayesian model is used in the sequence memorizer in \cite{cheng2014temporal}. Others try to model temporal dynamics with mid-level concept detectors \cite{bhattacharya2014recognition} or encounter the relation to speech processing and extract features comparable to acoustic features from a speech signal which are then fed into a classical HMM based speech recognizer \cite{chen2011modeling}.
Recently, a combination of CNNs with recurrent neural networks has successfully been applied to a fully supervised segmentation task \cite{yeung2016endtoend}. Richard and Gall \cite{richard2016temporal} propose to use a length model and a statistical language model in order to capture context information between different action classes. While their setup is specifically
designed for fully supervised learning, our system features weak supervision while still modeling context using a grammar and length using the HMM.

Additionally, \eg in the context of hierarchical action detection, where long activities can be subdivided into smaller action units, several works make use of grammars \cite{kuehne2014language,pirsiavash2014parsing,vo2014stochastic,kuehne2016end}. While the method of \cite{vo2014stochastic} is particularly designed for hierarchical action detection and makes use of a forward-backward algorithm over the grammar's syntax tree, our method builds on \cite{kuehne2016end}, where a HMM is used in combination with a context free grammar to parse the most likely video segmentation and corresponding action classification. While all previously mentioned methods use fully supervised training and rely on annotation of all actions and their exact segment boundaries, our method only requires the sequence of actions that occur in the video but no segment boundaries at all.

In terms of weakly supervised video segmentation and action detection, first attempts have been made by Laptev \etal \cite{laptev2008learning} using movie scripts for an automatic annotation of human actions in video. They align movie scripts with movie subtitles and use the temporal information associated with the subtitles to infer time intervals for a specific action. Building upon this work, Duchenne \etal \cite{duchenne2009automatic} also use an automatic script alignment to provide coarse temporal localization of human actions and point out that the textual based detection usually leads to temporal misalignment. Another approach has been made in \cite{sun2015temporal}. Here, the authors use only video-level annotation, \eg the main action class in the video, and try to find an appropriate localization of the actual actions using web images found under the name of the action class.
Most relevant to our method is the work \cite{bojanowski2014weakly} using the same kind of weak annotation that is also used in our approach and applying a discriminative clustering exploiting ordered action labels. In \cite{wu2015watch}, Wu \etal propose an algorithm to model long-term dependencies and co-occurrences of actions in a long activity via topic priors and time distributions between action-topic pairs. They feature a k-means based clustering to build action-words and model activities as sequences of these words. In \cite{huang2016connectionist}, a recurrent network has been deployed for weakly supervised learning. They extend a connectionist temporal classification (CTC) approach by a visual similarity measure between video frames to obtain reasonable alignments of frames to action labels. Similar to our approach, both the task of action alignment and joint classification and segmentation are tackled.

%Also related to our work, yet not solving the same problem, are some video-to-text methods that recently gained a lot of attention \cite{bojanowski2015weakly,venugopalan2015sequence,sener2015unsupervised,yao2015describing}. While some approaches require fully supervised training, such as the recurrent neural networks with attention mechanisms from \cite{yao2015describing} or the LSTM model of \cite{venugopalan2015sequence}, there are other weakly supervised techniques that make use of specific clustering techniques \cite{bojanowski2015weakly,sener2015unsupervised}. Particularly the work of \cite{bojanowski2015weakly} is of interest here, since it directly builds on \cite{bojanowski2014weakly} and generalizes their approach to handle continuous representations of text. They learn an alignment matrix that allows to reconstruct the segments and text labels. Note that this way of obtaining a segmentation differs from our approach to infer segment boundaries by aligning a HMM. We will later show that already \cite{bojanowski2014weakly}, which \cite{bojanowski2015weakly} is based on, fails on the complex datasets we use for evaluation.

\section{Transcript Annotation vs.\ Segment Annotation}

\begin{figure}[t]
\begin{center}
%a)\includegraphics[width=0.45\linewidth]{./Bild_0.png} 
%b)\includegraphics[width=0.45\linewidth]{./Bild_1.png} \\
a)\includegraphics[width=0.45\linewidth]{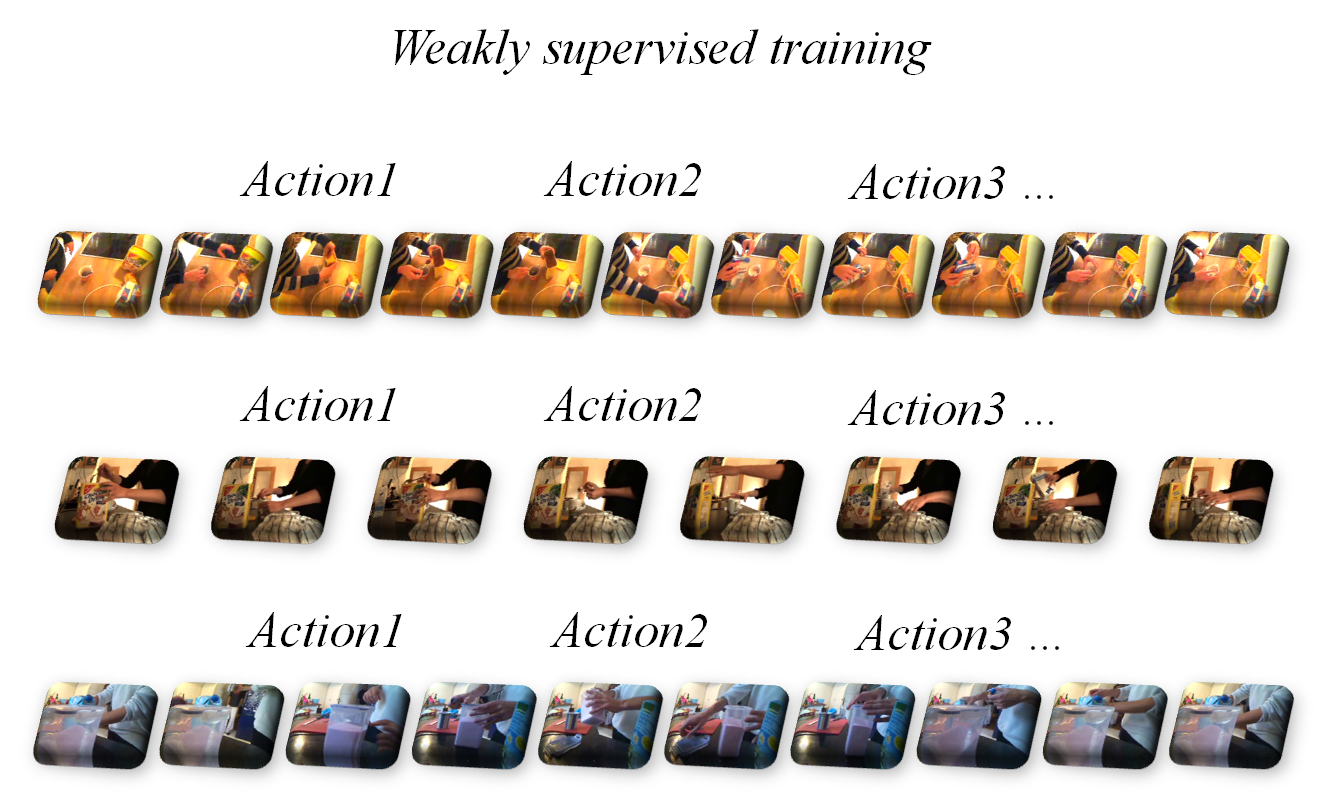} 
b)\includegraphics[width=0.45\linewidth]{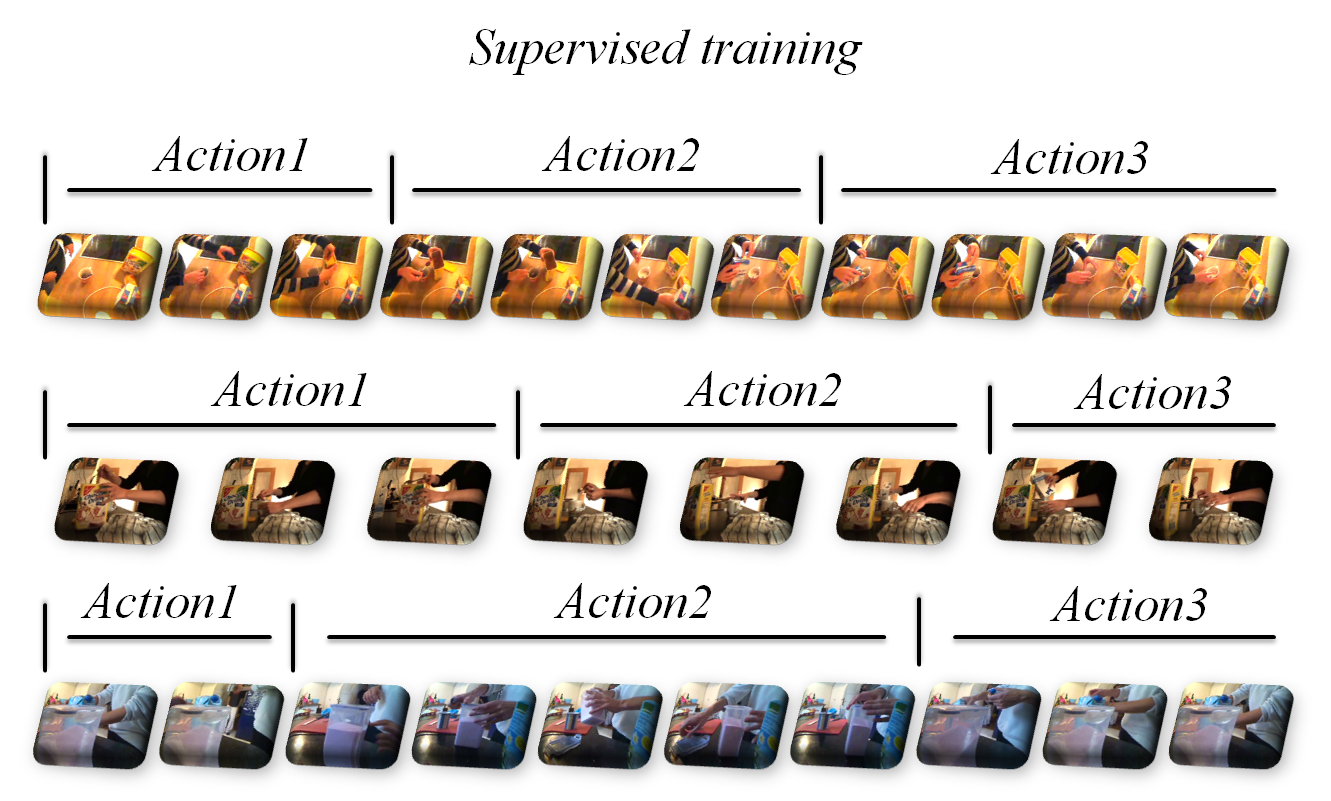} 
\end{center}
\caption{Problem statement of weakly supervised vs. supervised training. In case of weakly supervised training, transcript annotations define only the order in which the actions occur. A temporal labeling of frames is not given (a). In case of supervised training, segmentation information is provided with temporal information (b).  }
\label{fig:seg_vs_trans}
\end{figure}

Before we start with the actual system description, we describe the different types of annotation. 
We refer to the term \textit{transcripts} if the annotation contains only the actions within a video and the order in which they occur (Figure \ref{fig:seg_vs_trans} a), \eg 

\begin{center}
    \texttt{background, take\_cup, pour\_coffee, pour\_milk, background}.
\end{center}

Opposed to that, a fully segmented annotation, also referred to as segmentation, explicitly requires the start and end times or frames for each action (Figure \ref{fig:seg_vs_trans} b), \eg
\begin{center}
    \texttt{\phantom{00}0 - \phantom{0}61: background\phantom{0}} \\
    \texttt{\phantom{0}62 - 197: take\_cup\phantom{000}} \\
    \texttt{198 - 376: pour\_coffee} \\
    \texttt{277 - 753: pour\_milk\phantom{00}} \\
    \texttt{753 - 928: background\phantom{0}}. \\
\end{center}

To analyze the cost of the different annotation techniques, we let annotators label both types and compare the overall annotation time. We chose $ 10 $ videos from the Breakfast dataset with the activity `making coffee'. For this activity, there are seven possible action classes occurring within the videos. We let four test persons create a segmented annotation of the videos and measured the time they needed. Similarly, we let four other test persons create transcripts for the same videos and again measured the time. For both groups, we subtract the overall duration of the videos, and measure the additional time as annotation overhead. It shows that for a fully annotated segmentation annotators needed an overhead of $ 112.7 $ seconds on average to create the annotations for 10 videos. For the transcripts, annotators needed an overhead of only $ 14.1 $ seconds on average. Note that the difference can be expected to be even larger when the action segments in the video are shorter, e.g. as in the case of the fine-grained MPII dataset. Thus, being able to use weakly supervised data (\ie transcripts) is beneficial since the annotation is much cheaper. Additionally, transcripts may even be directly extracted from subtitles~\cite{laptev2008learning, alayrac2016unsupervised} without any annotation cost at all.

\section{System Overview}

In this work, we tackle the problem of video segmentation by assigning an action class to each frame.
To this end, we represent each action class by one HMM. Each HMM is defined by a number of states. Assigning a frame to a certain state also means that this frame belongs to the HMM to which the state belongs and to the action class that is represented by this HMM. Using this concept, we model a sequence of actions as a concatenation of HMMs. 

Inference, the association of each frame to a class, can be done by maximizing the probability for a given sequence of HMMs based on the observation probabilities for the single states. For observation probabilities, we consider two different computations. First, for the weak learning task, we consider GMMs to model observation probabilities of each state for a given input. We follow the approach of \cite{kuehne2016end} and model the observation probability of for each state by a single Gaussian component, but the described method generalizes for any number of mixtures.
Second, as an extension, we consider observation probabilities based on CNN output. As this requires a training of the respective CNN model, we perform this step as a post processing after the weak learning itself. More detailed, we use the boundaries inferred by the learning procedure and train a model based on those outputs.
In the following, we describe the components of the proposed architecture in detail.

\subsection{Action Model}
\label{sec:modelInitialization}

Based on the work of \cite{kuehne2014language}, we define a HMM for each action class that is defined by the set of states $ \mathcal{S} = \{ \varphi_1, \dots , \varphi_n \}$, the state transition probability matrix $ A \in \mathbb{R}^{n \times n}$ with the elements $a_{i,j}$ defining the transition from state $\varphi_i$ to state $\varphi_j$, and the observation probabilities $ b_j(x_t) = p(x_t | \varphi_j) $. We discuss these elements, the HMM topology and the generation of observation probabilities, in the following in detail.

\paragraph{Model topology} 
In our approach HMMs are defined by a strict left-to-right feed forward topology. So, only self-transitions and transitions to the next state are allowed. This means that within the state transition probability matrix $A$ only the elements $a_{i,i}$ and $a_{i,i+1}$ are nonzero.

\paragraph{Observation probabilities}
Additionally, we need to model the observation probabilities $p(x_t | \varphi_j)$ for being in a state $\varphi_j$ given an input sequence $\mathbf{x}_1^T$ and time $t$. More detailed, we denote the input sequence of each video as $ \mathbf{x}_1^T = \{ x_1, \dots , x_T \}$ and refer to $ x_t \in \mathbb{R}^m $ for the feature vector at frame $ t $.

To compute $p(x_t | \varphi_j)$, the observation probability is modeled by a multivariate Gaussian distribution defined as
\begin{align}\label{equ:pdf}
p(x_t|\varphi_j) = \frac{1}{\sqrt{(2\pi)^l|\Sigma_j|}}e^{-\frac{1}{2}(x_t-\mu_j)^\intercal\Sigma_j^{-1}(x_t-\mu_j)},
\end{align}
with $ l $ denoting the dimension of the input sequence $ \textbf{x} $,  $ \mu_j $ the $l$-dimensional mean vector, and $ \Sigma_j $ the $l \times l$ covariance matrix of the Gaussian model for state $ \varphi_j $. Since we model the observation probability by a single component Gaussian, the observation probability is parametrized only by the mean $ \mu_j $ and the covariance matrix $ \Sigma_j $ of the related feature distribution, which will allow for an easy update during the iterative optimization process.

\subsection{Inference}
\label{sec:inference}

\begin{figure}[t]
\begin{center}
%a)\includegraphics[width=0.45\linewidth]{./Bild_0.png} 
%b)\includegraphics[width=0.45\linewidth]{./Bild_1.png} \\
a)\includegraphics[width=0.45\linewidth]{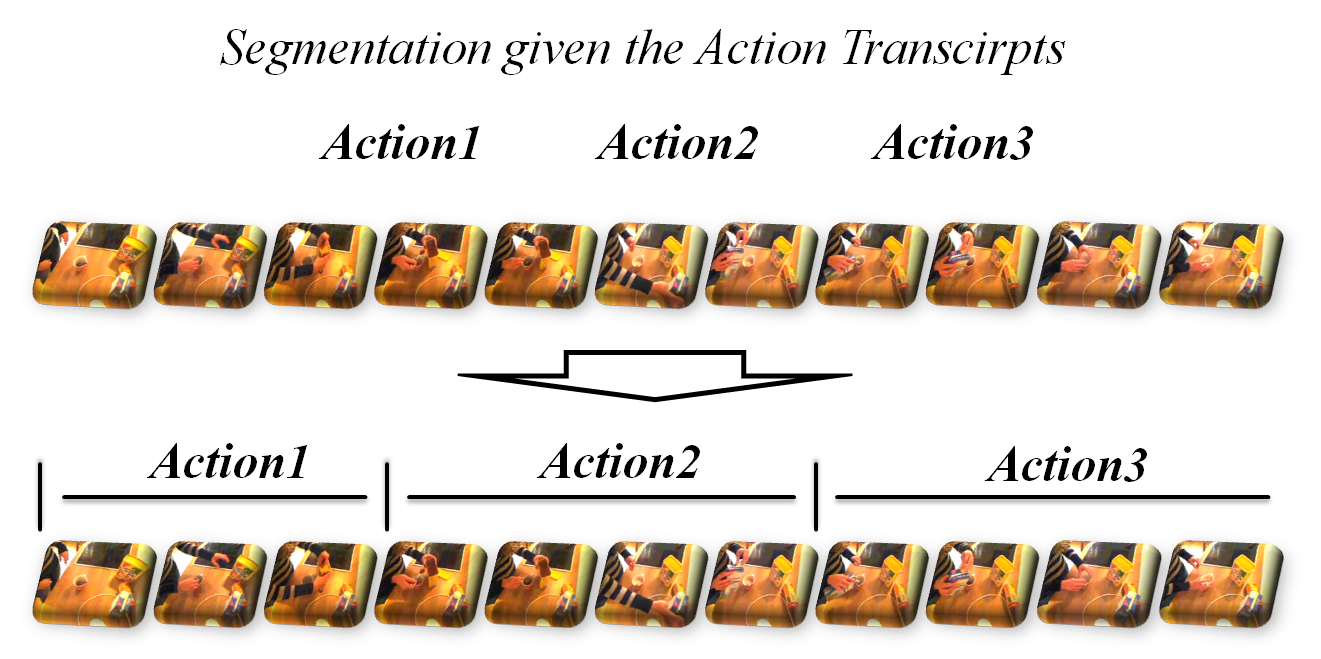} 
b)\includegraphics[width=0.45\linewidth]{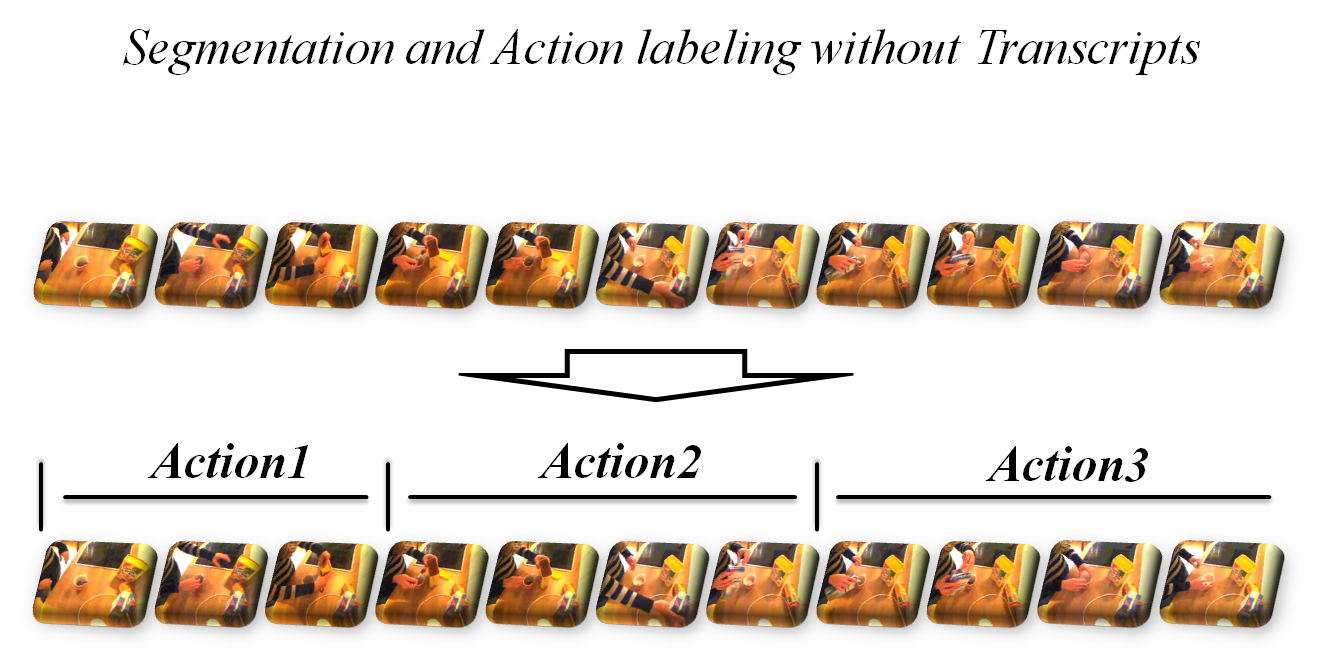} \\
\end{center}
\caption{Example for video segmentation given the action transcripts compared to combined segmentation and action labeling without transcripts. In the first case the order of actions is given by the transcripts and only segmentation boundaries need to be inferred (a). In the second case no transcripts are provided and both, the sequence of occurring actions and the segment boundaries, are hypothesized for each action (b). }
\label{fig:seg_vs_class}
\end{figure}

In this section, we describe how to infer the segmentation of a video by finding the best alignment of video frames to a sequence of one or more HMMs. We consider two different cases: First we describe inference for the case that the transcripts are provided and we are interested in the segmentation boundaries (Figure \ref{fig:seg_vs_class} a). We refer to this task as alignment of video with transcripts or, shortly, alignment task. This case is used in training as well as for the segmentation, when given a list of transcripts. Second, we describe inference for the case that no transcripts are provided and we need to estimate both, the sequence of occurring actions as well as the segment boundaries for each hypothesized action (Figure \ref{fig:seg_vs_class} b). This case is used for the combined classification and segmentation of video sequences. 

\paragraph{Video Alignment given the Action Transcripts}

Given the action transcripts, it is possible to concatenate the HMM representation of multiple actions in the order they occur in the transcript for this sequence into one large sequence-HMM. Video segmentation is done by finding the best alignment of video frames to the sequence-HMM. The most likely alignment of the input frames to the sequence-HMM states is
\begin{align}
    \argmax_{s_1,\dots,s_T} \big\{ p(s_1,\dots,s_T|\mathbf{x}_1^T) \big\} = \argmax_{s_1,\dots,s_T} \big\{ \prod_{t=1}^T p(x_t|s_t) \cdot p(s_t|s_{t-1}) \big\},
    \label{equ:argmax}
\end{align}
where each $ s_t \in \mathcal{S} $ is a state of one of the original action class HMMs. Hence, $ s_1,\dots,s_T $ assigns a HMM state to each frame of the input sequence. The above maximization can be solved efficiently using the Viterbi algorithm. Once the alignment from frames to HMM states is computed, it is straightforward to infer the segment boundaries by looking at which points $ t $ there is a transition from the HMM of one action class to the HMM of another action class. An example for the boundary information obtained by applying the initialized models to the video sequence input is e.g. shown in Figure \ref{fig:exp_iterations_reestimation} b.

\paragraph{Video Segmentation and Action Labeling without Transcripts}

If only the video but not its action transcripts are given, we need to infer both, the segment boundaries and the actual actions that occur in the video. A general approach is to define all valid action label sequences with a context free grammar and evaluate all possible paths. Since the typical datasets are frequently rather short and there is only a limited amount of training sequences, we use a minimalist grammar by considering all possible paths that occur in the training data. More precisely, let $ \varphi_1^{(n)},\dots,\varphi_K^{(n)} $ be the sequence-HMM of the $ n $-th training sequence. The grammar $ \mathcal{G}_n $ generates all alignments $ s_1,\dots,s_T $ of length $ T $ to this sequence-HMM. As there is only a finite number of possible alignments, $ \mathcal{G}_n $ is finite and thus context free. Now let $ \mathcal{G} $ be the union of all $ \mathcal{G}_n $. Then, $ \mathcal{G} $ is a context free grammar generating all possible alignments of length $ T $ that match a sequence-HMM from the training data. For inference, only these alignments have to be considered:
\begin{align}
    \argmax_{s_1,\dots,s_T \in \mathcal{G}} \big\{ p(s_1,\dots,s_T|\mathbf{x}_1^T) \big\} = \argmax_{s_1,\dots,s_T \in \mathcal{G}} \big\{ \prod_{t=1}^T p(x_t|s_t) \cdot p(s_t|s_{t-1}) \big\}.
\end{align}
Note that although the amount of possible alignments $ s_1,\dots,s_T \in \mathcal{G} $ is exponential in the number $ T $ of frames, it is possible to infer the optimal alignment in polynomial time using a Viterbi algorithm.

\subsection{Weakly Supervised Learning}
\label{sec:weakSupTraining}

So far, we have introduced the model and showed how to use it for inference. It remains to show how it can be trained using the action transcripts only.

To this end, we compute the probability of an alignment going through state $ \varphi_j $ at time $ t $, $ p(s_t=\varphi_j | \mathbf{x}_1^T) $,
which can be computed as the product of a forward and backward probability $ \alpha_j(t) $ and $ \beta_j(t) $,
\begin{align}
    \alpha_j(t) &= p(s_t=\varphi_j|\mathbf{x}_1^t), \\
    \beta_j(t)  &= p(s_t=\varphi_j|\mathbf{x}_t^T).
\end{align}

The probabilities $ \alpha_j(t) $ and $ \beta_j(t) $ are computed using the Baum-Welch algorithm.
This information is then used to update the single HMM models independently. We calculate the forward probabilities $\alpha_j(t)$ and backward probabilities $\beta_j(t)$ for all states $ \varphi_j $ and times $t$. For each state $ \varphi_j $ and time $t$, we use the product of the two probabilities and the current observation vector to update the multivariate Gaussian distribution, namely the mean $ \mu_j $ and covariance matrix $ \Sigma_j $ (Eq.~\eqref{equ:pdf}),  for this state. Each observation is assigned to every state in proportion to the probability of the model being in that state when the vector was observed. The resulting new estimates are

\begin{align}
    \mu_j' =  \frac{\sum_{t=1}^T \alpha_j(t) \beta_j(t) x_t}{\sum_{t=1}^T \alpha_j(t) \beta_j(t)},
\end{align}
\begin{align}
    \Sigma_j' =  \frac{\sum_{t=1}^T \alpha_j(t) \beta_j(t) (x_t - \mu_j)(x_t - \mu_j)^T}{\sum_{t=1}^T \alpha_j(t) \beta_j(t)}.
\end{align}

\begin{figure}[t]
\begin{center}
\includegraphics[width=0.9\linewidth]{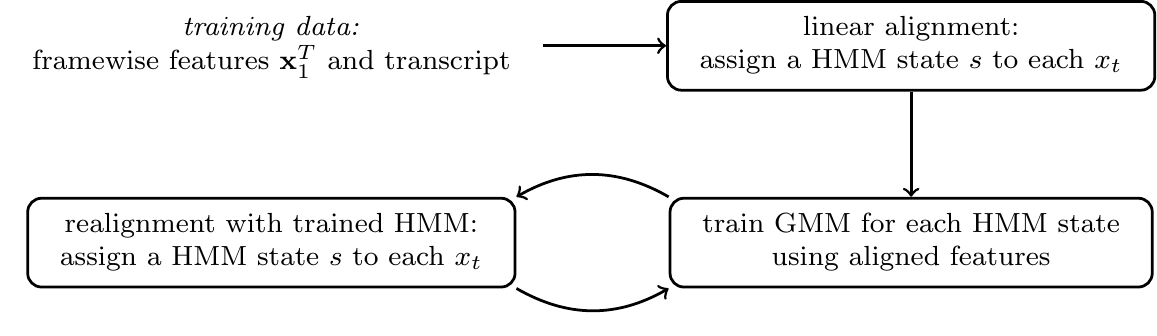} 
\end{center}
\caption{Weakly supervised training procedure with action transcripts. At the beginning, the HMMs of each action instance in the video are concatenated and the frames are linearly distributed over the HMM states. This initial alignment is used to train the GMMs and obtain a HMM that better fits the training data. Using this HMM, a realignment of frames to HMM states is computed. The procedure is iterated until convergence.}
\label{fig:hmmTraining}
\end{figure}

\begin{figure}[t]
\begin{center}
\includegraphics[width=0.9\linewidth]{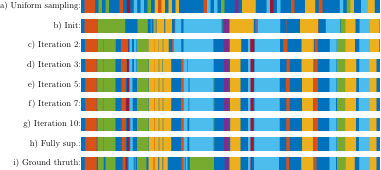} 
\end{center}
\caption{This example segmentation of one video from the MPII Cooking dataset shows how the segmentation evolves during the training. We start with a uniform sampling to initialize the model (a) and apply the initialized models to the overall sequence (b). The new boundaries are then used to train the new model parameters and the procedure is repeated until convergence (c-g). After three to five iterations, the inferred segmentation is fairly similar to the fully supervised segmentation (h) as well as to the annotated ground truth (i).}
\label{fig:exp_iterations_reestimation}
\end{figure}

With the updated model parameters, a new alignment can be computed as described in Section \ref{sec:inference}. With this new alignment, the model parameters can again be updated. This process is iterated until convergence.  An overview of the process is shown in Figure \ref{fig:hmmTraining}. An example for the segmentation results after each step is shown in Figure \ref{fig:exp_iterations_reestimation}.

Note that the initialization can be seen as a crucial factor for the overall system. If the linear alignment of frames to HMM states hardly matches the true action segments at all, \eg if the segment lengths are very unbalanced, the system might learn an unrealistic segmentation. However, if at least a certain amount of frames of the desired action is included in each initial segment, the algorithm usually converges quickly and achieves satisfying results. Also note that not all actions need to have a good initial model, as long as the overall parameters can be inferred by the subsequent Viterbi decoding.

\subsection{Extension to CNN models}
\label{sec:extFullySup}

After a reasonable good segmentation has been found by the proposed weak learning method, the existing model can easily be extended by more powerful models such as CNNs. 
To this end, we exploit the fact that the softmax layer of a CNN generates a posterior distribution over all output classes, e.g. given the sequence input $\mathbf{x}$ at frame $t$ the output for the class $s$ is $p(s|x_t)$. As the proposed action deals with conditional probabilities we can transform the softmax-layer output by using the Bayes` rule
\begin{align}
    p(x_t|s) = \frac{p(s|x_t)}{p(s)} \cdot p(x_t) .
\end{align}
The class prior probability $ p(s) $ is estimated as the relative frequency of the state $ s $ in the current state-to-frame alignment of the training data.
The factor $ p(x_t) $ can be omitted as it does not affect the maximizing arguments in Equation~\eqref{equ:argmax}.
In the training, all states of all HMMs are treated as independent classes. For each frame $t$, the CNNs will thus produce probabilities for all possible HMMs states, which corresponds to the observation probability matrix $b_j$ and can thus be handled in the same manner as the output of the multivariate Gaussian distribution for the generic features.
Note that for the CNN-based system, only the GMM is replaced by a CNN and the remainder of the system stays unchanged.

\section{Experiments}

In this section, we evaluate our method empirically for the two setups described in Section \ref{sec:inference}, the segmentation task, where we only infer the segmentation of a video given the transcripts, as well as the combined segmentation and classification, where the segmentation and action labels are inferred simultaneously without transcript information. In this case, the transcripts are only used during training to learn the models and build a grammar for the recognition system.

\subsection{Datasets}

%\paragraph{Datasets.}
Our method is evaluated on the following four action detection datasets:

\paragraph{Hollywood Extended} \cite{bojanowski2014weakly} is a dataset of $ 937 $  clips extracted from Hollywood movies. It features $ 15 $ action classes with a mean of $ 2.5 $ segments per video. For evaluation, we use ten-fold cross-validation, using the last digit in the file name as respective split index.

\paragraph{MPII Cooking} \cite{rohrbach2012database} is a large database for
the detection of fine grained cooking activities. It comprises $ 8 $h of video data with recordings of $ 12 $ different individuals. Including background, the dataset consists of $ 65 $ action classes with a mean of $ 86.1 $ segments per video. We use seven-fold cross-validation with the same splits that are also used in \cite{rohrbach2012database}.

\paragraph{Breakfast} \cite{kuehne2014language} is a large scale database allowing for hierarchical activity recognition and detection and comprises roughly $ 77 $h of video and about $ 4 $ million frames. A set of $ 10 $ breakfast related activities is divided into $ 48 $ smaller action classes with $ 4.9 $ segments per video. We use the fine grained action annotation to evaluate the segmentation accuracy of our method. Recognition results on the $ 10 $ coarse activity classes are also reported. Following \cite{kuehne2014language}, we use four splits for evaluation.

\paragraph{CRIM13} \cite{Burgos12CRIM} is a large-scale mice behavior dataset, featuring about $ 50 $h of annotated mice activities, capturing the interactions of two mice as well as isolated behaviour. Overall 13 different action classes are annotated, with about 140 action segments within ten minutes of footage. The dataset comprises the side as well as the top view of a transparent cage. For the following evaluation, we only consider the side view of all activities. Additionally, as mice behaviour is clearly different from goal directed human behaviour in the other datasets, we favour a bi-gram model for sequence parsing instead of a fixed path grammar.

\subsection{Experimental Setup}

\subsubsection{Generic Feature Computation}

For the proposed system, we use a combination of dense trajectories and reduced Fisher vectors. We compute dense trajectory features \cite{wang2013action} and reduce the dimensionality of the descriptor from $ 426$  dimensions to $ 64 $ dimensions by PCA, as described in~\cite{oneata2013action}. 
To compute the Fisher vectors, we follow the protocol of \cite{kuehne2016end} and sample 100,000 random features to learn a Gaussian mixture model with 256 GMMs. The Fisher vector representation is computed for each frame over a sliding window of 20 frames using the implementation of~\cite{vedaldi2008vlfeat}. The dimensionality of the resulting vector is then reduced to 64 dimensions using PCA again. Thus, each frame is represented by a 64-dimensional reduced Fisher vector. We further apply an L2-normalization to each feature dimension separately for each video clip to further normalize the features. 

\subsubsection{Model Initialization}
To build a model for each action, we first need to define the number of states for each model. The number of states for each HMM is set to a fraction of the mean action length in frames. The mean action length is computed based on the transcripts, \ie it is obtained by dividing the number of frames by the number of action instances in the transcripts. We chose the number of states such that each state captures $ 10 $ video frames on average, respectively. E.g. in case of Breakfast, 29 states per HMM were used, resulting in an overall of 1392 states, assigned to 48 classes, for this dataset.

Second we need to initialize the state transition probability matrix $A$. The state transition probabilities $ a_{i,j} $ of each HMM are initialized based on the average number of frames per state. As a state captures $ 10 $ frames on average, we set the transition probability $ a_{i,i+1} $ to $ 1/10 $ and the self-transition probability $ a_{i,i} $ to $ 9/10 $. As we feature a left-to-right feed forward model, all other probabilities are zero.

\subsubsection{Initialization of observation probabilities}
The observation probability parameters mean $ \mu $ and the covariance matrix $ \Sigma $ are initialized using a linear alignment of the action transcript to the video frames. Each video sequence is split into $ k $ segments of equal size, where $ k $ is the number of action instances in the transcript (see e.g. Figure \ref{fig:exp_iterations_reestimation} a). This way, we obtain an initial alignment of video frames to HMMs that can be used to train the Gaussian mixture models that specify the distributions $ b_j(x) = p(x|s_j) $.

\subsubsection{CNN training}

For the here proposed approach we use a two-stream architecture \cite{wang15towards, simonyan2014two, feichtenhofer2016conv}, as this has shown to give good results in context of action recognition. Because of the extensive and time consuming training procedure, we evaluated the extension with CNN features for two out of four datasets, the Hollywood extended and the Breakfast dataset.

We follow the training and evaluation protocol as described in \cite{wang15towards}, using single RGB images to train the spatial stream and 10 frames of optical flow in $x$- and $y$-direction to train the temporal stream. For each split of each dataset, we fine-tune a CNN with the respective state classes, 238 for Hollywood extended and 1392 for Breakfast, as output. We use a VGG architecture that has been initialized for UCF101 as provided by \cite{wang15towards} and run it with a batch-size of 50 frames for 10,000 iteration for the spatial stream and for 30,000 iterations for the temporal stream without additional data augmentation for training or testing. 
As the datasets are very heterogeneous in terms of spatial and temporal variety and some tend to overfit more easily then other, the recognition accuracy varied with the number of iterations. To find the best configuration, we computed the average performance for 10K, 15K, 20K, 25K, and 30K iterations for each split on a separate validation set and used the best configuration for testing. 

For testing, we compute probabilities for every frame and use those for inference as described in Section \ref{sec:extFullySup}. Overall, we found that for all tested datasets the spatial stream produces consistently worse results than the temporal stream, and also the combination of both inputs does not outperform the performance of the temporal stream alone. This holds especially for datasets like Breakfast, where the background does not provide any additional context information in contrast to datasets like UCF101. We therefore use only the output of the temporal stream in all experiments.

\subsection{Aligning Videos to Action Transcripts}
\label{sec:segmentationFromTranscripts}

We first look at the alignment capabilities of our model. In this case transcripts are provided and the aim is to infer the start and end frames of the respective action units.  We investigate the quality of the hypothesized segments on both, training data and test data. In both cases the action transcripts are given, but we only use the training data to learn the parameters of the model.

We provide four performance measures for all experiments: We report mean over frames (MoF), which is the mean frame wise accuracy of the hypothesized video alignment, and mean over classes (MoC), computed by taking the mean over frames for each class independently and computing the average over all classes. We also report the Jaccard index (Jacc) by computing over all frames the intersection over union (IoU) of ground truth and recognized action segment defined by $|G \cap D|/|G \cup D|$ with $G$ referring to the ground truth frames and $D$ referring to the detected action frames. Additionally, we include the Jaccard index as intersection over detection (IoD) as proposed by \cite{bojanowski2014weakly}, computed by intersection of correctly detected action segments and all detected action segments $|G \cap D|/|D|$. Both measures are also reported as mean over all classes.

Looking at the characteristics of the provided measures, mean over frames weights each frame equally, so frequent or long action instances have a higher impact on the result. This is particularly important if one class dominates a dataset, which is often the case for the background class. Recognizing this class correctly can already lead to high MoF rates. For mean over classes, in contrast, this effect is averaged out. In exchange, rare classes may have an unreasonable large influence to the MoC score since each class contributes equally.

\begin{table}[t] \footnotesize
    \centering
    \begin{tabularx}{\textwidth}{llp{0.1cm}Y|YYYY|Y}
        \toprule
        & & & \multicolumn{6}{c}{Alignment on training set for iterations 3, 5, and 10}  \\
              \cmidrule{4-9}             
        Dataset & & & naive & init & 3rd  & 5th  &  10th  & fully \\
        \midrule\rule{0pt}{3ex}
        \multirow{4}{*}{HwdExt}
        & \textit{MoF} &       & $ 0.472 $ & $ 0.451 $ & $ 0.494 $ & $ 0.508 $ & $ \textbf{0.516} $ & $ 0.691 $  \\
        & \textit{MoC} &       & $ 0.367 $ & $ 0.426 $ & $ \textbf{0.483} $ & $ 0.471 $ & $ 0.466 $ & $ 0.723 $  \\
        & \textit{Jacc(IoU)} & & $ 0.221 $ & $ 0.263 $ & $ \textbf{0.291} $ & $ 0.290 $ & $ 0.290 $ & $ 0.544 $  \\ 
        & \textit{Jacc(IoD)} & & $ 0.417 $ & $ 0.446 $ & $ 0.469 $ & $ 0.478 $ & $ \mathbf{0.482} $ & $ 0.675 $  \\ 
        \midrule
        \multirow{4}{*}{MPII}
        & \textit{MoF} &       & $ 0.192 $ & $ 0.455 $ & $ 0.577 $ & $ 0.584 $ & $ \textbf{0.590} $ & $ 0.863 $  \\
        & \textit{MoC} &       & $ 0.128 $ & $ 0.385 $ & $ 0.492 $ & $ 0.498 $ & $ \textbf{0.499} $ & $ 0.910 $  \\
        & \textit{Jacc(IoU)} & & $ 0.068 $ & $ 0.228 $ & $ 0.318 $ & $ 0.324 $ & $ \textbf{0.327} $ & $ 0.792 $  \\
        & \textit{Jacc(IoD)} & & $ 0.144 $ & $ 0.341 $ & $ 0.430 $ & $ 0.440 $ & $ \mathbf{0.446} $ & $ 0.850 $  \\ 
        \midrule
        \multirow{4}{*}{Breakfast}
        & \textit{MoF} &       & $ 0.312 $ & $ 0.393 $ & $ 0.439 $ & $ \textbf{0.440} $ & $ 0.434 $ & $ 0.879 $  \\
        & \textit{MoC} &       & $ 0.281 $ & $ 0.355 $ & $ \textbf{0.404} $ & $ 0.401 $ & $ 0.392 $ & $ 0.883 $  \\
        & \textit{Jacc(IoU)} & & $ 0.174 $ & $ 0.228 $ & $ \textbf{0.266} $ & $ 0.265 $ & $ 0.261 $ & $ 0.782 $  \\
        & \textit{Jacc(IoD)} & & $ 0.347 $ & $ 0.401 $ & $ \mathbf{0.426} $ & $ \mathbf{0.426} $ & $ 0.425 $ & $ 0.863 $  \\ 
        \midrule
        \multirow{4}{*}{CRIM13}
        & \textit{MoF} & & $ 0.306 $ & $ 0.298 $ & $ 0.380 $ & $ 0.394 $ & $ \textbf{0.401} $ & $ 0.665 $  \\
        & \textit{MoC} & & $ 0.155 $ & $ 0.289 $ & $ 0.361 $ & $ 0.377 $ & $ \textbf{0.383} $ & $ 0.692 $  \\
        & \textit{Jacc(IoU)} & & $ 0.083 $ & $ 0.131 $ & $ 0.179 $ & $ 0.189 $ & $ \textbf{0.193} $ & $ 0.454 $  \\
        & \textit{Jacc(IoD)} & & $ 0.150 $ & $ 0.210 $ & $ 0.271 $ & $ 0.287 $ & $ \mathbf{0.292} $ & $ 0.571 $  \\ 
        \bottomrule
    \end{tabularx}
    \caption{Comparison of the video alignment quality on the training data for naive uniform alignment (\textit{naive}), initialization (\textit{init}),  weakly supervised training (\textit{3rd, 5th and 10th iteration}), and fully supervised training (\textit{fully}). Results are reported as mean over frames (\textit{MoF}), mean over classes (\textit{MoC}) and Jaccard index as intersection over union (\textit{Jacc}(\textit{IoU})) and over detection (\textit{Jacc}(\textit{IoD})).} % In case of training, more iterations obviously help to adapt the model to the training data,  with best results reached for 10 iterations, whereas in case of test data best results are most often reached after three iterations, indicating an overfitting on the training data after three iterations. 
    \label{tab:alignment_train}
\end{table}

 \begin{table}[t] \footnotesize
    \centering
   \begin{tabularx}{\textwidth}{llp{0.1cm}Y|YYYY|Y}
        \toprule
        & & & \multicolumn{6}{c}{Alignment on test set for iterations 3, 5, and 10}   \\
              \cmidrule{4-9}           
        Dataset & & & naive & init & 3rd  & 5th  &  10th  & fully  \\
        \midrule\rule{0pt}{3ex}
        \multirow{4}{*}{HwdExt}
        & \textit{MoF} &       & $ 0.473 $ & $ 0.441 $ & $ 0.491 $ & $ 0.505 $ & $ \textbf{0.513} $ & $ 0.645 $  \\
        & \textit{MoC} &       & $ 0.350 $ & $ 0.419 $ & $ \textbf{0.476} $ & $ 0.465 $ & $ 0.457 $ & $ 0.617 $  \\
        & \textit{Jacc(IoU)} & & $ 0.217 $ & $ 0.267 $ & $  0.302 $ & $ \textbf{0.305} $ & $ 0.298 $ & $ 0.457 $  \\
        & \textit{Jacc(IoD)} & & $ 0.414 $ & $ 0.436 $ & $  0.460 $ & $ \mathbf{0.468} $ & $ 0.463 $ & $ 0.593 $  \\ 
        \midrule
        \multirow{4}{*}{MPII}
        & \textit{MoF} &       & $ 0.182 $ & $  0.463 $ & $ \textbf{0.598} $ & $ 0.596 $ & $ 0.560 $ & $ 0.662 $  \\
        & \textit{MoC} &       & $ 0.109 $ & $ 0.354 $ & $ \textbf{0.436} $ & $ 0.426 $ & $ 0.427 $ & $ 0.503 $  \\
        & \textit{Jacc(IoU)} & & $ 0.057 $ & $ 0.220 $ & $ 0.309 $ & $ 0.312 $ & $ \textbf{0.314} $ & $ 0.408 $  \\
        & \textit{Jacc(IoD)} & & $ 0.119  $ & $ 0.386  $ & $ 0.490 $ & $ \mathbf{0.523} $ & $ 0.517 $ & $ 0.690 $  \\ 
        \midrule
        \multirow{4}{*}{Breakfast}
        & \textit{MoF} & & $ 0.304 $ & $ 0.374 $ & $ \textbf{0.400} $ & $ 0.386 $ & $ 0.376 $ & $ 0.738 $  \\
        & \textit{MoC} & & $ 0.245 $ & $ 0.330 $ & $ \textbf{0.347} $ & $ 0.319 $ & $ 0.302 $ & $ 0.567 $  \\
        & \textit{Jacc(IoU)} & & $ 0.165 $ & $ 0.223 $ & $ \textbf{0.244} $ & $ 0.230 $ & $ 0.221 $ & $ 0.452 $  \\
        & \textit{Jacc(IoD)} & & $ 0.327 $ & $ 0.381 $ & $ \mathbf{0.406} $ & $ 0.384 $ & $ 0.379 $ & $  0.667 $  \\ 
        \midrule
        \multirow{4}{*}{CRIM13}
        & \textit{MoF} & & $ 0.304 $ & $ 0.298 $ & $ 0.383 $ & $ 0.391 $ & $ \textbf{0.392} $ & $ 0.646 $  \\
        & \textit{MoC} & & $ 0.153 $ & $ 0.287 $ & $ \textbf{0.357} $ & $ 0.352 $ & $ 0.342 $ & $ 0.643 $  \\
        & \textit{Jacc(IoU)} & & $ 0.081 $ & $ 0.137 $ & $ \textbf{0.187} $ & $ 0.186 $ & $ 0.181 $ & $ 0.436 $  \\
        & \textit{Jacc(IoD)} & & $ 0.147 $ & $ 0.230 $ & $ 0.303 $ & $ \mathbf{0.318} $ & $ 0.316 $ & $ 0.577 $  \\ 
        \bottomrule
    \end{tabularx}
    \caption{Comparison of the video alignment quality on the test data for naive uniform alignment (\textit{naive}), initialization (\textit{init}),  weakly supervised training (\textit{3rd, 5th and 10th iteration}), and fully supervised training (\textit{fully}). \textit{Test} refers to the alignment results on unseen data. } % In case of training, more iterations obviously help to adapt the model to the training data,  with best results reached for 10 iterations, whereas in case of test data best results are most often reached after three iterations, indicating an overfitting on the training data after three iterations. 
    \label{tab:alignment_test}
\end{table}

In Table \ref{tab:alignment_train} and \ref{tab:alignment_test}, we evaluate the alignment performance on the training as well as on the test data. First column, \textit{naive}, is the naive baseline of each dataset. Here, the video is simply split into $ k $ segments of equal size, where $ k $ is the number of action instances in the transcripts. We assign the labels of the action transcripts to each corresponding segment as shown in Figure \ref{fig:exp_iterations_reestimation} a and compute the alignment quality with respect to the ground truth. We can see that for all datasets the naive baseline already classifies 20\%-30\% of all frames correctly. 
As this splitting is used to initialize the related action models, this is also the amount of correct frames the system starts with. 

The second column, \textit{init}, shows the result of the initialized HMMs applied for alignment before the first realignment (see also Figure \ref{fig:exp_iterations_reestimation} b). One can see that this first inference based on the initialized models and transcripts leads to a significant improvement in alignment accuracy. This shows that the linear alignment is enough to estimate the model parameters and can be seen as a hint that the initialization is successful. The new boundaries are then used for further reestimation. 

For the \textit{weakly} supervised results, we report results after three, five and ten iterations (Figure \ref{fig:exp_iterations_reestimation} d, e and g). It shows that the accuracy increases significantly compared to the initial model especially for the first iterations.
Comparing the results for the train and the test data, it can be observed that the systems starts to converge after three to five iterations and begins to overfit on the training data leading to a decreased accuracy in the test data.

\begin{figure}[t]
\begin{center}
    \small
	\includegraphics[width=0.45\linewidth]{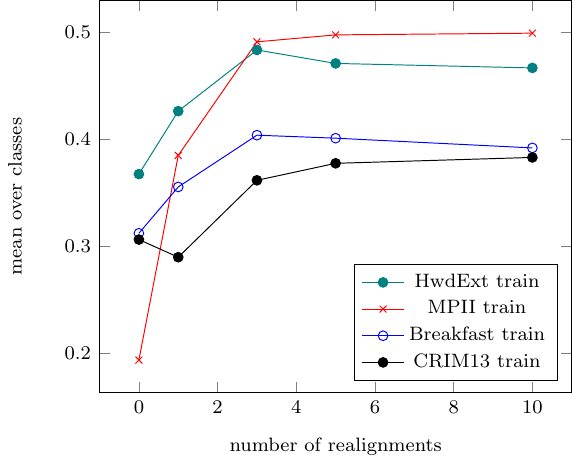} 
    \hspace{5mm}
	\includegraphics[width=0.45\linewidth]{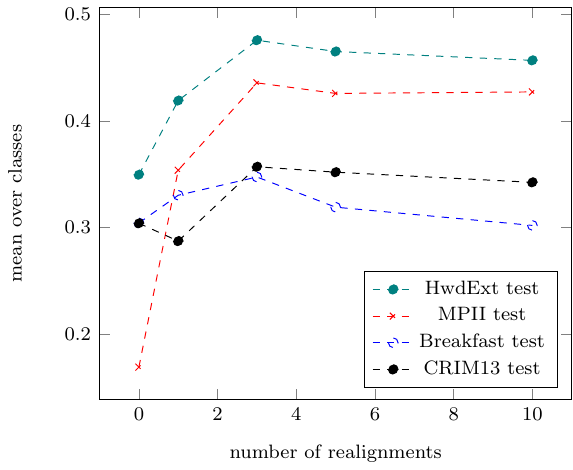} 
\end{center}
\caption{Effect of the number of realignments on the alignment results on train and test set of Hollywood Extended, MPII, Breakfast, and CRIM13 (MoC).}
\label{fig:runs}
\end{figure}

To further analyze the effect of the number of iterations we plot alignment results for all four datasets for up to 10 iterations in Figure \ref{fig:runs}. On all datasets, the obtained alignment on both train and test set improves until the third iteration. After that, a degradation of the performance can be observed on the test data. We therefore fix the number of iterations for all following experiment to three iterations.

% evaluation of subsampling

\begin{figure}[t]
\begin{center}
\includegraphics[width=0.9\linewidth]{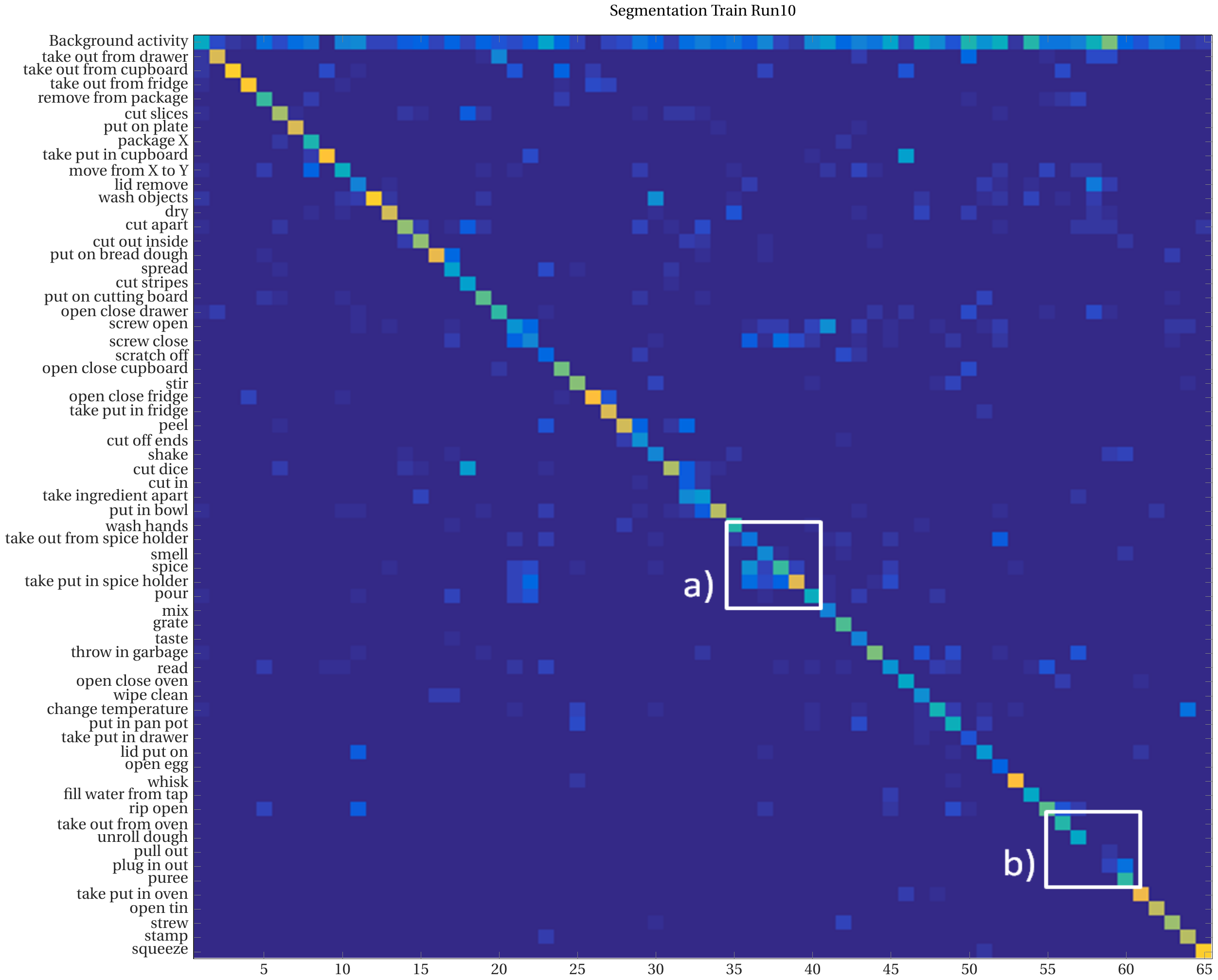} 
\end{center}
\caption{Confusion matrix for alignment on MPII test set after ten iterations. Especially classes that often appear in successive order tend to be aggregated, such as ``take out from spice holder'',  ``smell'', ``spice'', ``take put in spice holder'' (a).  Additionally, classes with minor instances tend to be suppressed (b). }
\label{fig:exp_MPII_seg}
\end{figure}

Looking at the confusion matrix after ten iterations as shown in Figure \ref{fig:exp_MPII_seg} for MPII, it shows two main reasons for this overfitting. First, successive classes that consistently appear in the same order are aggregated, here e.g. ``take out from spice holder'',  ``smell'', ``spice'', ``take put in spice holder''. This is based on the fact that, if a certain combination of actions is always executed in the same order and the classes do not appear in another context, segment boundaries can be set at any point within those segments without decreasing the probability of the overall sequence. 
Second, the accuracy for classes with few instances, such as ``pull out'' decreases. All datasets used for evaluation have a highly imbalanced distribution of class instances, \eg for MPII the number of instances of the largest and the smallest differ by a factor of $ 42.8 $. Thus, classes with few samples will be initialized with very few data, and in the following only be detected for a few frames. Therefore, those models tend to degenerate during training. Accordingly, classes with more training samples will be recognized more often and thus result in more general models such as e.g. the background class.

We also compare the results to the \textit{fully} supervised setup (Table \ref{tab:alignment_train} and  \ref{tab:alignment_test}). Here we use the ground truth for the initial assignment of frames to HMM states (Figure \ref{fig:exp_iterations_reestimation} h) and do not apply any boundary reestimation. It shows that on test data the weakly learned models catch up \eg to 5\%-10\% on Hollywood Extended and MPII compared to the fully supervised approach. Further, it achieves at least 50\%, and usually more accuracy compared to the fully supervised models. 

\begin{figure}
    \centering
    \subfloat[alignment of the video \texttt{s16$-$d11$-$cam002}.]{\includegraphics[scale=0.28]{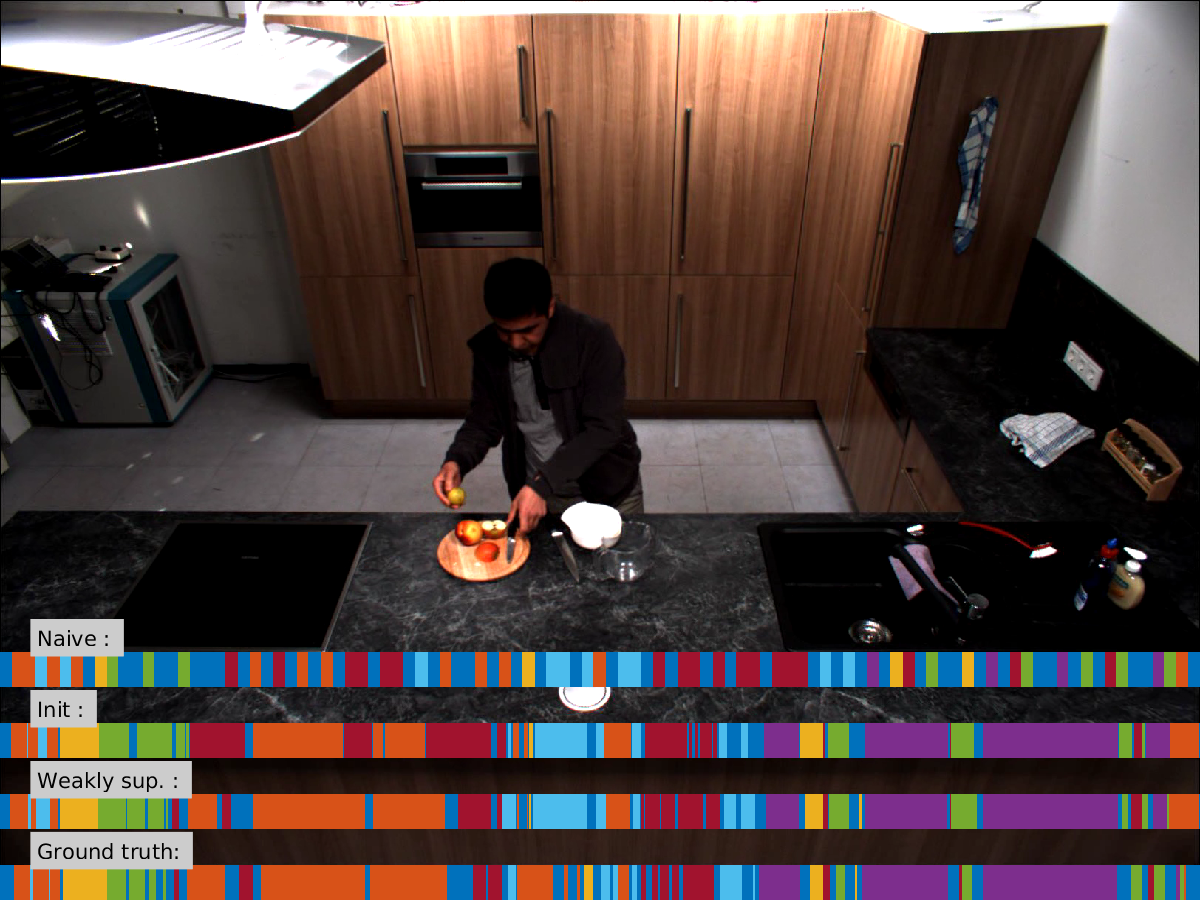}}
    \hfill
    %\subfloat[alignment of the video \texttt{s16$-$d06$-$cam002}.]{\includegraphics[scale=0.28]{./s16-d06-cam-002_frame700_b.png}}\\
    \subfloat[alignment of the video \texttt{P15\_cereal}.]{\includegraphics[scale=0.28]{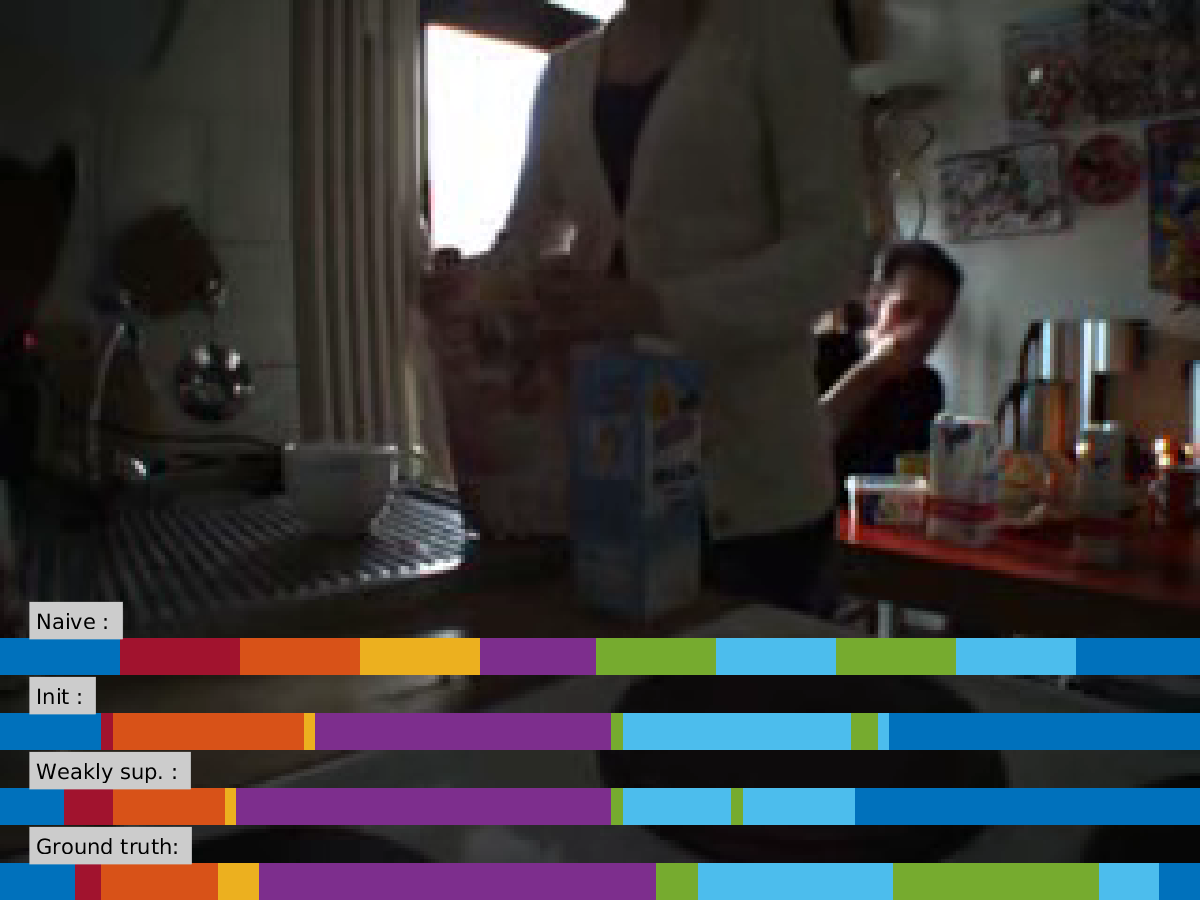}}
    %\hfill    
    %\subfloat[alignment of the video \texttt{P13\_cereal}.]{\includegraphics[scale=0.28]{./image_0200_3.png}}
    \caption{Example for video alignment from action transcripts, showing first the naive baseline, the alignment after the initialization, after three iterations as well as the ground truth segmentation. It shows that the first inference after the initialization already leads to an acceptable alignment, that is further refined during training. }
    \label{fig:exp_seg}
\end{figure}

Figure \ref{fig:exp_seg} shows qualitative alignment results for videos from MPII Cooking and Breakfast dataset. The first color bar shows the \textit{naive} alignment. The second color bar, \textit{init}, refers to the inference result of initial model. The third and fourth color bar show the alignment output and the annotated ground truth respectively.

\subsection{Extension to CNN models}

As shown before, we fixed the number of iterations for all experiments to three iterations. To extend the framework by CNN features we thus use the alignment generated on the training set for each split to train the CNN. We report results for the case of using only CNN features, as well as for the case of combined CNN and GMM features. In the later case, we just take the mean over both state probabilities. Overall it shows that, for the alignment task, the CNN features are able to outperform the generic features in some cases, but not consistently (see Table \ref{tab:alignment_test_CNN}). Namely in case of Hollywood Extended, which is the smaller of both datasets, they are performing worse than the generic features, whereas in case of Breakfast, where more training data is available, they significantly improve the alignment. It also shows that, for both dataset, the CNN features are able to improve the mean over class accuracy. Looking at the results in detail, it shows that they are able to classify especially classes with few instances better than the GMM based features. It also shows, that in case of Breakfast the combination of GMM and CNN performs slightly better than both features alone.  

 \begin{table}[t] \footnotesize
    \centering
   \begin{tabularx}{\textwidth}{llp{0.1cm}YYY}
        \toprule
        & & & \multicolumn{3}{c}{Alignment with generic and CNN features}   \\
              \cmidrule{3-6}           
        Dataset & & & GMM  & CNN alone  &  GMM + CNN    \\
        \midrule\rule{0pt}{3ex}
        \multirow{4}{*}{HwdExt}
        & \textit{MoF}       & &  $ \mathbf{0.491} $ & $ 0.479 $ & $ 0.489 $  \\
        & \textit{MoC}       & &  $ 0.476 $ & $ \mathbf{0.527} $ & $ 0.481 $  \\
        & \textit{Jacc(IoU)} & &  $ \mathbf{0.302} $ & $ 0.296 $ & $ 0.294 $  \\
        & \textit{Jacc(IoD)} & &  $ \mathbf{0.460} $ & $ 0.407 $ & $ 0.430 $  \\
        \midrule
        %\rule{0pt}{3ex} 
        \multirow{4}{*}{Breakfast}
        & \textit{MoF}       & &  $ 0.400 $ & $ 0.471 $  & $ \mathbf{0.473} $   \\
        & \textit{MoC}       & &  $ 0.347 $ & $ \mathbf{0.536} $  & $ 0.530 $   \\
        & \textit{Jacc(IoU)} & &  $ 0.244 $ & $ 0.308 $  & $ \mathbf{0.313} $   \\
        & \textit{Jacc(IoD)} & &  $ 0.406 $ & $ 0.451 $  & $ \mathbf{0.460} $   \\
        \bottomrule
    \end{tabularx}
    \caption{Comparison of the video alignment quality on the test data for weakly supervised training with generic features (\textit{GMM}), fine-tuned CNN features (\textit{CNN}) and the combination of both (\textit{GMM + CNN}). } % In case of training, more iterations obviously help to adapt the model to the training data,  with best results reached for 10 iterations, whereas in case of test data best results are most often reached after three iterations, indicating an overfitting on the training data after three iterations. 
    \label{tab:alignment_test_CNN}
\end{table}

\subsection{Segmentation from Video Data Only}
\label{sec:segmentvideo}

In this section, we evaluate the resulting models of the weak learning for the combined classification and segmentation of a video sequence, \ie localizing and classifying the actions. Here video features of the test set are used without any further information. The task is then to localize the respective action classes and to segment them just based on the models gained by weak learning. 
To this end, we again compare three different training modalities: (a) for the initial model based on linear segmentation only, (b) after the weakly supervised training, and (c) for the fully supervised training in Table \ref{tab:segmentation2}. Note that in this case, no annotations are used at all for the parsing of the test sequences. This is different from the segmentation task in Section \ref{sec:segmentationFromTranscripts}, where we infer the segmentation given the action transcripts for the test data. In order to model the temporal dependencies between the hypothesized actions, a context free grammar is used, \cf Section \ref{sec:inference}. We again report  mean over frames (\textit{MoF}), mean over classes (\textit{MoC}) and Jaccard index (Jacc) as intersection over union. The Jaccard index as intersection over detection is only meaningful for the alignment task, \ie when the action transcript is given during inference. Thus, we omit this metric for the segmentation experiments. For the Breakfast dataset, we also include \textit{Activity}, which is the high level activity classification accuracy, using the context free grammar to look up which high level activity the recognized action sequence belongs to.  
\begin{table}[tb]
    \centering
    \footnotesize
    \begin{tabularx}{\textwidth}{llXcXcXc}
        \toprule
        & & \multicolumn{6}{c}{Segmentation with initial|weakly|fully supervised model }  \\
              \cmidrule{3-8}             
        Dataset      & & & init & & weakly sup. & & fully sup.  \\
        \midrule\rule{0pt}{3ex}
        \multirow{3}{*}{HwdExt}
        & \textit{MoF} &  & $ 0.259 $   & & $ 0.330 $   & & $ 0.395 $   \\
        & \textit{MoC} &  & $ 0.109 $   & & $ 0.186 $   & & $ 0.177 $   \\
        & \textit{Jacc(IoU)} & & $ 0.042 $   & & $ 0.086 $   & & $ 0.084 $   \\
        \midrule
        %\rule{0pt}{3ex} 
        \multirow{3}{*}{MPII}
        & \textit{MoF}  & & $ 0.443 $   & & $ 0.597 $   & & $ 0.720 $   \\
        & \textit{MoC}  & & $ 0.329 $   & & $ 0.432 $   & & $ 0.572 $   \\
        & \textit{Jacc(IoU)} & & $ 0.196 $   & & $ 0.297 $   & & $ 0.455 $   \\
        \midrule
        %\rule{0pt}{3ex} 
        \multirow{4}{*}{Breakfast}
        & \textit{MoF}  & & $ 0.163 $   & & $ 0.259 $   & & $ 0.507 $   \\
        & \textit{MoC}  & & $ 0.123 $   & & $ 0.167 $   & & $ 0.327 $   \\
        & \textit{Jacc(IoU)} & & $ 0.052 $   & & $ 0.098 $   & & $ 0.361 $   \\
        & \textit{Activity} & & $ 0.405 $   & & $ 0.566 $ & & $ 0.664 $   \\
        \midrule
        %\rule{0pt}{3ex} 
        \multirow{3}{*}{CRIM13}
        & \textit{MoF} &  & $ 0.113 $   & & $ 0.238 $   & & $ 0.328 $   \\
        & \textit{MoC} &  & $ 0.191 $   & & $ 0.287 $   & & $ 0.530 $   \\
        & \textit{Jacc(IoU)} & & $ 0.053 $   & & $ 0.108 $   & & $ 0.187 $   \\
        \bottomrule
    \end{tabularx}
    \caption{Comparison of the combined video classification and segmentation quality for the initial model as well as the weakly supervised model (three iterations) and fully supervised model, respectively. Results are reported in mean  over frames (\textit{MoF}), mean over classes (\textit{MoC}) and Jaccard index (\textit{Jacc}(\textit{IoU})). \textit{Activity} is the high level activity classification accuracy on Breakfast.}
    \label{tab:segmentation2}
\end{table}
Naturally, applying fully supervised training with all data achieves better results than just weakly supervised training.
Still, it becomes clear that the gap between both training methods is in some cases comparably small, e.g. for mean over classes on Hollywood Extended or MPII Cooking. Also for the task of activity recognition on the Breakfast dataset, weakly supervised training achieves competitive results compared to fully supervised training.

\begin{figure}
    \centering
    \subfloat[Classification and segmentation result of the video \texttt{s16-d01-cam002}.]{\includegraphics[scale=0.28]{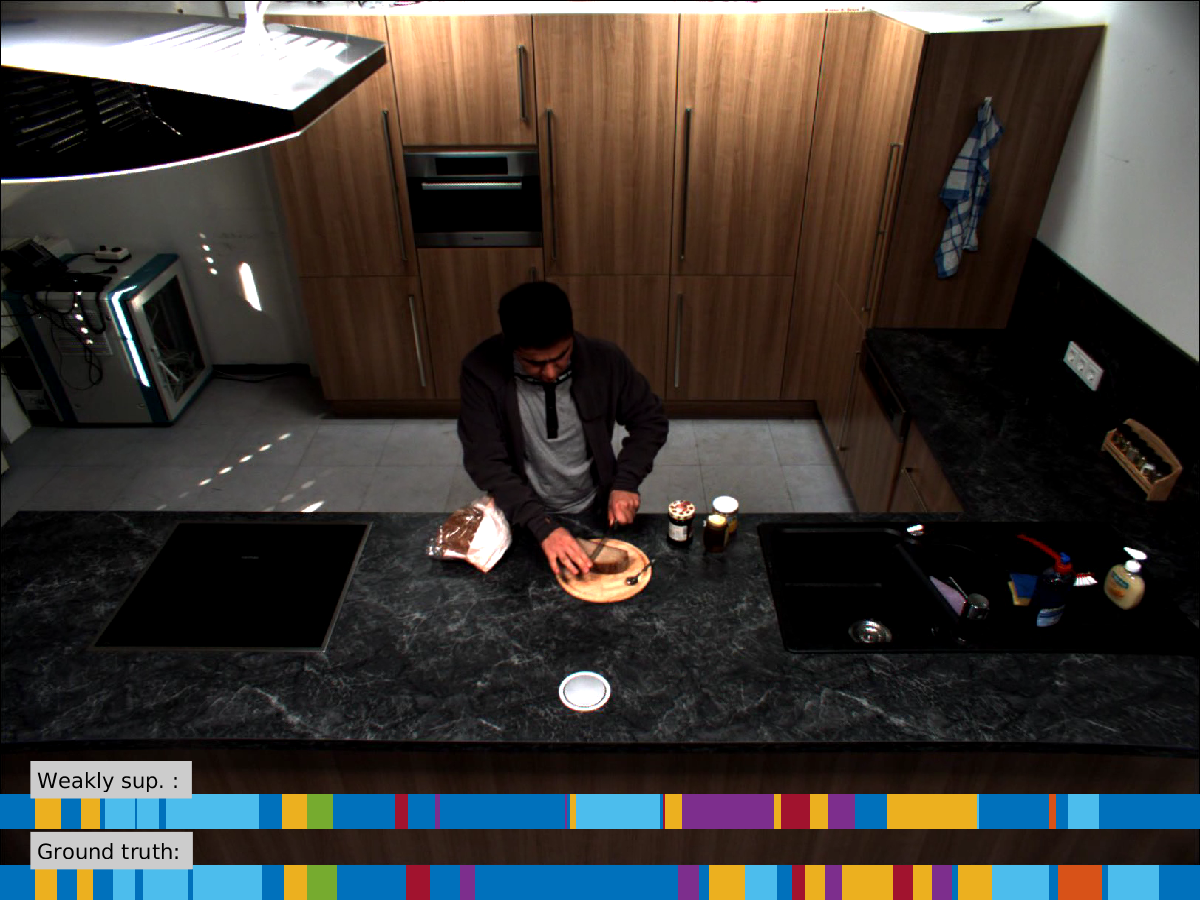}}
    \hfill
    %\subfloat[Classification and segmentation result of the video \texttt{s16-d09-cam002}.]{\includegraphics[scale=0.28]{./s16-d09-cam-002_frame1700_b.png}}\\
    \subfloat[Classification and segmentation result of the video \texttt{P12\_sandwich}.]{\includegraphics[scale=0.28]{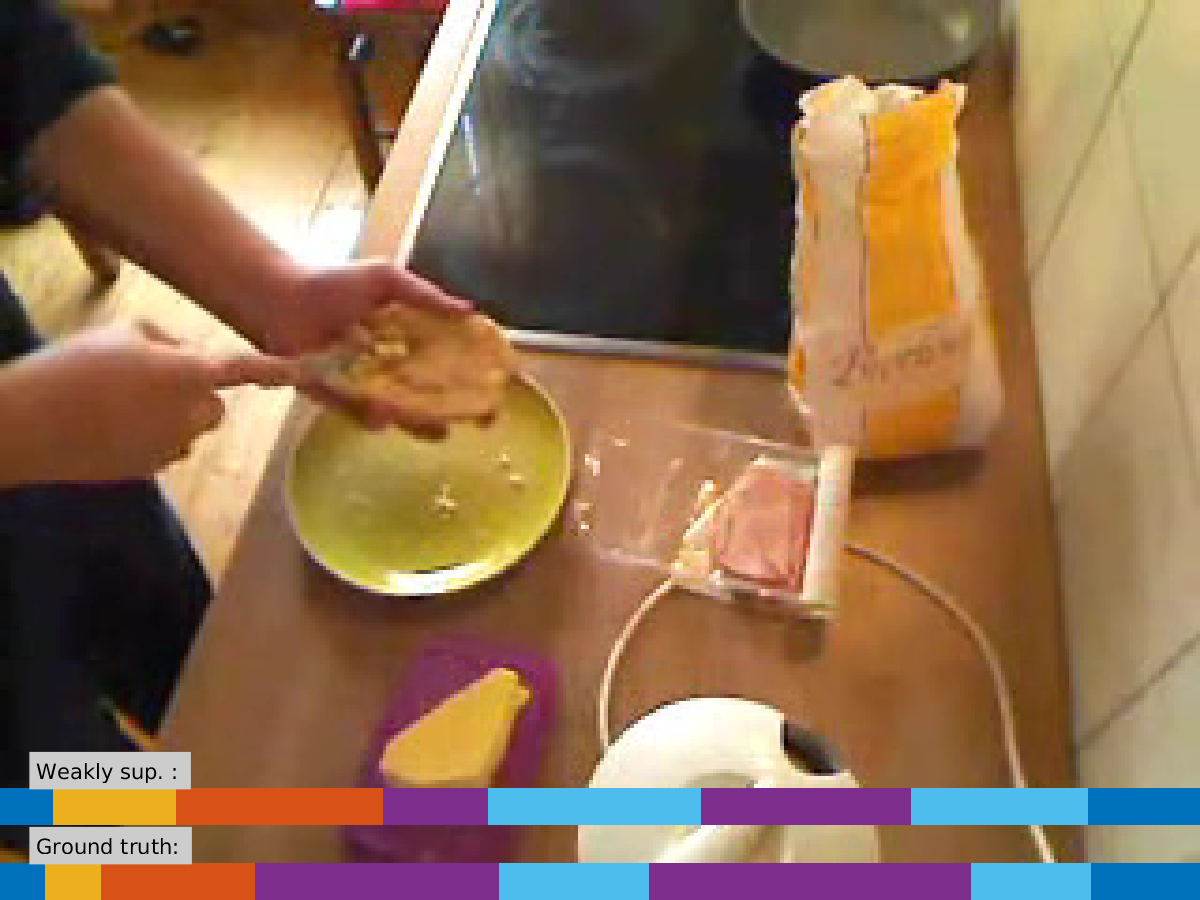}}
    %\hfill
    %\subfloat[Classification and segmentation result of the video \texttt{P08\_milk}.]{\includegraphics[scale=0.28]{./image_0700.png}}
    \caption{Example for the combined classification and segmentation without transcript information. It shows that the inferred results deviate more from the ground truth compared to the results of the segmentation task. }
    \label{fig:exp_seg_class}
\end{figure}

Figure \ref{fig:exp_seg_class} shows example results for the combined segmentation and classification on MPII Cooking and Breakfast. Since we do not use the transcripts in this setup, the result deviates more from the from the ground truth compared to the results of the segmentation task shown in Figure \ref{fig:exp_seg}.   
 \begin{table}[t] \footnotesize
    \centering
   \begin{tabularx}{\textwidth}{llp{0.1cm}YYY}
        \toprule
        & & & \multicolumn{3}{c}{Segmentation with generic and CNN features}   \\
              \cmidrule{3-6}           
        Dataset & & & GMM  & CNN alone  &  GMM + CNN    \\
        \midrule\rule{0pt}{3ex}
        \multirow{3}{*}{HwdExt}
        & \textit{MoF}       & & $ \mathbf{0.330} $ & $ 0.287 $ & $ 0.245 $  \\
        & \textit{MoC}       & & $ 0.186 $ & $ 0.246 $ & $ \mathbf{0.298} $  \\
        & \textit{Jacc(IoU)} & & $ 0.086 $ & $ 0.103 $ & $ \mathbf{0.106} $  \\
        \midrule
        %\rule{0pt}{3ex} 
        \multirow{4}{*}{Breakfast}
        & \textit{MoF}       & & $ 0.259 $ & $ 0.263 $ & $ \mathbf{0.282} $   \\
        & \textit{MoC}       & & $ 0.167 $ & $ \mathbf{0.248} $ & $ 0.219 $   \\
        & \textit{Jacc(IoU)} & & $ 0.098 $ & $ 0.116 $ & $ \mathbf{0.129} $   \\
        & \textit{Activity}  & & $ 0.566 $ & $ 0.629 $ & $ \mathbf{0.653} $   \\
        \bottomrule
    \end{tabularx}
    \caption{Comparison of the combined video classification and segmentation quality for the initial model as well as the weakly supervised model (three iterations) and fully supervised model, respectively. Results are reported in mean  over frames (\textit{MoF}), mean over classes (\textit{MoC}) and Jaccard index (\textit{Jacc}(\textit{IoU})). \textit{Activity} is the high level activity classification accuracy on Breakfast.}
    \label{tab:segmentation_GMM_CNN}
\end{table}

Considering the extension by CNN features for the task of segmentation in Table \ref{tab:segmentation_GMM_CNN}, it shows that the improvement becomes more clear than in the alignment task. Here in almost all cases, the CNN performs better than the GMM. Again, in case of Breakfast the improvement is more prominent than in case of Hollywood Extended and the most significant gain for both datasets is reached for the mean over class criterion. It becomes also visible, that the combination of both usually helps to improve the results even further.

\subsection{Comparison to State-of-the-art}

We compare the proposed framework to the system of Bojanowski \etal \cite{bojanowski2014weakly}, which is also able to infer video segmentation boundaries given weak annotations as well as to the approach by Huang \etal \cite{huang2016connectionist} based on connectionist temporal models. For our system, we fix the number of iterations for all datasets to three. We consider the alignment as well as the segmentation task and report Jaccard index as intersection over detection (\textit{Jacc}(\textit{IoD})) for the alignment and mean over frames (\textit{MoF}) for the segmentation task. The results are shown in Table~\ref{tab:sota}.

For the alignment task, our best method (GMM on Hollywood Extended and GMM + CNN on Breakfast) outperforms existing approaches by $ 2\% $ to $ 5\% $. The performance boosts
over ECTC is particularly remarkable, as the CTC criterion is specifically designed to align sequences to input frames. Note that the choice of the best system, \ie GMM or GMM + CNN, again depends on the amount of available training data.

For the segmentation task, we can only compare to related work on the Breakfast dataset since neither OCDC nor ECTC have been applied on Hollywood Extended. The authors of \cite{huang2016connectionist} report $ 8.9\% $  MoF on Breakfast for the OCDC \cite{bojanowski2014weakly} and $ 27.7\% $ for the ECTC approach, which is slightly better than our system with GMMs. However, including CNN features boosts our system by $ 2.5\% $, closing the gap to ECTC. Eventually, our system outperforms current state-of-the-art on both, the alignment and the segmentation task.
 
\begin{table}
    \centering
    \footnotesize
    \begin{tabularx}{\textwidth}{llYYYY}
        \toprule
        & & GMM & GMM+CNN   & OCDC \cite{bojanowski2014weakly} & ECTC \cite{huang2016connectionist} \\
        \midrule
        \textit{Alignment Task} \\
        HwdExt & \textit{Jacc(IoD)}    & $ \mathbf{0.460} $      & $ 0.430 $         & $ 0.234 $         & $ 0.410 $     \\
        Breakfast & \textit{Jacc(IoD)} & $ 0.406 $      & $ \mathbf{0.460} $         & $ 0.439 $         & -             \\
        \midrule
        \textit{Segmentation Task} \\
        Breakfast & \textit{(MoF)}     & $ 0.259 $      & $ \mathbf{0.282} $         & $ 0.089 $         & $ 0.277 $     \\
        \bottomrule
    \end{tabularx}
    \caption{Comparison of our model (GMM only and GMM + CNN) with other weakly supervised methods, OCDC~\cite{bojanowski2014weakly} (evaluated by \cite{huang2016connectionist}) and extended CTC~\cite{huang2016connectionist}. Only for the alignment task, the transcripts are given during inference.}
    \label{tab:sota}
\end{table}

\subsection{Runtime}

Finally, we look at the runtime of the weak learning process. We report the runtime for training with three iterations, for the alignment on the test data as well as for combined classification and segmentation, see Table~\ref{tab:runtime}. We process the first split of each dataset on one core of an Intel® Core™ i7 CPU with 3.30GHz and compared it with computation time of \cite{bojanowski2014weakly} under same conditions. The reported time refers only to the inference resp. optimization process and does not include feature computation or loading. Note that segmentation refers to the combined classification and segmentation without any transcripts as described in Section~\ref{sec:segmentvideo} as opposed to the alignment task with given transcripts (Section~ \ref{sec:segmentationFromTranscripts}).

\begin{table}[tb]
    \centering
    \footnotesize
    \begin{tabularx}{\textwidth}{XYYYY}
        \toprule
         Dataset  &  Train & Alignment & Segmentation & \cite{bojanowski2014weakly} -Train  \\
        \midrule
         HwdExt & $ 0.47 $ min & $ 0.06 $ min   & $ 1.61 $ min   & $ 8.2  $ min    \\
         MPII & $ 5.1 $ min & $ 0.93 $ min   & $ 13.7 $ min   & $ 42.6 $ min   \\
         Breakfast & $3.9$ min & $  0.35 $ min    & $  23.7 $ min   & $ 115.4  $  min  \\
        \bottomrule
    \end{tabularx}
    \caption{Evaluation of runtime for the weak learning with three iterations, alignment on the test data and combined classification and segmentation.}
    \label{tab:runtime}
\end{table}

Overall the proposed method is roughly one order of magnitude faster for training than \cite{bojanowski2014weakly}. This is mainly based on the different number of iterations required for both methods. As discussed in Section~\ref{sec:segmentationFromTranscripts}, we fixed the number of iterations to three, as the proposed method usually converges and starts to overfit at this point. For \cite{bojanowski2014weakly}, the authors recommend 200 iterations resulting thus in a longer overall runtime.

\section{Conclusion}

We proposed an approach for weakly supervised learning of a temporal action model. For training we use a combination of a frame-based video representation and the corresponding transcripts to infer the scripted actions, and thus, learn the related action models without the need of a frame level annotation. 
To do this, we model each action by a HMM and iterate sequence decoding and model reestimation to adapt and train the related models, based on the transcribed training data.
We evaluate our approach on four challenging activity segmentation datasets and showed that the process iteratively improves the estimation of the segment boundaries and the action classification. The weakly supervised learned action models are competitive in comparison to the models learned with full supervision, showing that weakly supervised learning for temporal semantic video segmentation is also feasible for large-scale video datasets.

\section{Acknowledgments}
The work has been financially supported by the DFG projects KU 3396/2-1 (Hierarchical Models for Action Recognition and Analysis in Video Data) and GA 1927/4-1 (DFG Research Unit FOR 2535 Anticipating Human Behavior) and the ERC Starting Grant ARCA (677650). This work has been supported by the AWS Cloud Credits for Research program.

\bibliography{paper_for_review}

\end{document}